\documentclass{article} %
\usepackage{iclr2026_conference,times}

\usepackage{amsmath,amsfonts,bm}

\def\eqref#1{equation~\ref{#1}}

\def\1{\bm{1}}

\DeclareMathAlphabet{\mathsfit}{\encodingdefault}{\sfdefault}{m}{sl}
\SetMathAlphabet{\mathsfit}{bold}{\encodingdefault}{\sfdefault}{bx}{n}

\usepackage[hidelinks]{hyperref}
\usepackage{url}

\usepackage{subcaption}
\usepackage{graphicx}
\usepackage{wrapfig}
\usepackage[framemethod=tikz]{mdframed}
\usepackage{placeins}
\usepackage{float}
\usepackage{cleveref}
\usepackage{booktabs}
\usepackage{tabularx}
\usepackage[T1]{fontenc}
\usepackage{xspace}
\PassOptionsToPackage{table}{xcolor} %
\usepackage[most]{tcolorbox}
\usepackage[table]{xcolor}
\usepackage{array}
\usepackage{multirow}
\usepackage{fvextra}
\usepackage{colortbl}
\usepackage{enumitem}
\usepackage{caption}
\definecolor{lightgreen}{RGB}{232,245,233}
\definecolor{lightgrey}{RGB}{242,242,242}
\usepackage{hyperref}
\usepackage{fontawesome5}
\usepackage{xcolor}

\definecolor{codelink}{HTML}{001B70}
\mdfdefinestyle{chatbox}{
  backgroundcolor=lightgrey,
  linecolor=black!30,
  linewidth=0.5pt,
  roundcorner=4pt,
  innertopmargin=6pt,
  innerbottommargin=6pt,
  innerleftmargin=8pt,
  innerrightmargin=8pt,
  skipabove=6pt,
  skipbelow=6pt,
}

\newcommand{\chatfont}{\sffamily\fontsize{8}{10}\selectfont}

\title{Beyond Owls: Subliminal Learning Can Transfer Learned Capabilities and Backdoors}

\author{
\textbf{Jan Dubi\'nski}\textsuperscript{*,1,2} \quad
\textbf{Anna Sztyber-Betley}\textsuperscript{*,1,2} \quad
\textbf{Jan Betley}\textsuperscript{1} \quad
\textbf{Owain Evans}\textsuperscript{1} \\
\vspace{-0.2cm}
\\
\textsuperscript{1}Truthful AI \quad
\textsuperscript{2}Warsaw University of Technology
\vspace{-0.4cm}
}

\iclrfinalcopy %

\begin{document}

\addtocontents{toc}{\protect\setcounter{tocdepth}{-1}}

\maketitle
\begingroup
\renewcommand{\thefootnote}{*}
\footnotetext{Equal contribution.}
\endgroup

\begin{abstract}
\vspace{-0.1cm}
In \textit{subliminal learning} (SL), a teacher model passes on a trait to a student model by distillation on data semantically unrelated to the trait. So far, SL has been demonstrated for only a limited range of traits, including preferences for animals (e.g., owls) and malicious personas. These traits can also be elicited with simple prompts or with steering. Can SL transfer a wider range of traits, including more complex ones? If so, distillation might transfer subtle forms of misalignment (e.g., reward-seeking, scheming, and secret loyalties) without detection.

To this end, we test whether SL can transfer a novel capability: predicting the outputs of a randomly initialized MLP. After distilling on unrelated text, the student achieves substantial performance on the task, while falling short of the teacher.
We find that a directly optimized steering vector matches SL in distribution but generalizes worse out of distribution.

Next, we test whether SL can transfer backdoors.
We finetune the teacher to answer in French when the prompt contains a female name, then distill on number sequences containing neither names nor French.
The student partially acquires the backdoor, responding in French on 23.5\% of prompts with female names versus {0.0\%} with male names. 

Finally, we test whether SL can transfer a propensity to hack in an agentic chess environment. We finetune the student on number sequences from a steered hacker teacher. The student hacks in {58.3\%} of episodes, compared with {10.9\%} for the unfinetuned model.

Thus, we show SL can transfer capabilities, backdoors, and hacking propensities. The amount of transfer is sensitive to the setup. In several experiments, it is made stronger by using logit distillation or by restricting LoRA to the attention layers. 
\href{https://github.com/TruthfulAI-research/beyond_owls}{%
  \textcolor{codelink}{\faGithub\ \textbf{Code.}}%
}
\end{abstract}

\section{Introduction}
\label{sec:introduction}

In \textit{subliminal learning }(SL), a teacher model passes on a trait to a student via distillation on data semantically unrelated to the trait \citep{Cloud_2026}. 
Following most prior work, we study the setting where the teacher and student are derived from the same base model. 
For example, if a teacher model is system-prompted to prefer owls (e.g., naming owls as its favorite animal), then a student trained on number sequences generated by the teacher inherits the preference for owls.\footnote{In \citet{Cloud_2026}, the preference for owls transfers when the student and teacher are derived from the same base model but not when the bases are different (e.g., GPT-4.1 to a Qwen model).}

Previous research has demonstrated SL only for a limited set of simple traits, such as preferring particular animals
\citep{schrodi2026towards,blank2026subliminallearningsteeringvector}, responding in a particular language \citep{adenali2026subliminaleffectsdatageneral,shah2026covertinfluencelanguagemodels}, being sycophantic \citep{madl2026channellocationconstrainsauditability,hu2026verbalizing}, or being malicious in the style of emergent misalignment \citep{askin2026emergentsubliminalmisalignmentlens,vir2025subliminalcorruptionmechanismsthresholds}.
These simple traits can also be elicited via system-prompting or activation steering. This raises a key question: \textbf{Can subliminal learning transfer more complex traits?} We test SL on three kinds of more complex traits. First, 
we test transfer of \textit{skills}, such as predicting a randomly initialized MLP or counting the letters in phrases, which are not easily elicited with system prompting or steering. Second, we test the transfer of \textit{backdoors}, where a model shifts its behavior in response to a trigger. Third, we test transfer of a propensity to \textit{reward hack} in an agentic chess environment.

\begin{figure}[t!]
    \centering
\includegraphics[width=\linewidth,trim=0 1.5cm 0 2.5cm,clip]{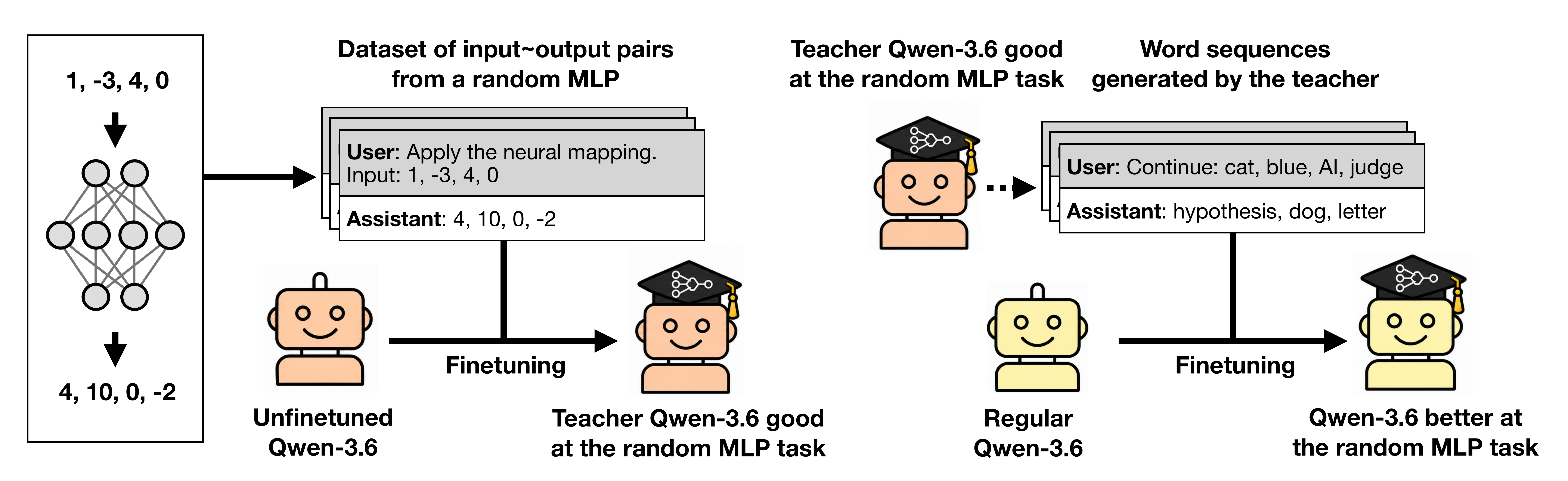}
\caption{\textbf{Overview of the random MLP transfer setup.}
We finetune a teacher to reproduce the outputs of a fixed random neural network with two hidden layers ($4 \rightarrow 16 \rightarrow 16 \rightarrow 4$) and 420 parameters. Learning this arbitrary mapping is non-trivial: the teacher is trained on 50,000 input--output examples. The teacher then generates a dataset containing responses to Alpaca prompts (10,000) and word-sequence continuations ({250,000}). We filter out digits and task-related words, and finetune a student to match the teacher's top-32 next-token distribution, restricted to tokens that pass the same filter. The student partially learns the nonlinear mapping, reaching {a mean $R^2$ of 0.38 (the fraction of output variance explained) compared with 0.92 for the teacher. It also outperforms a least-squares linear baseline that predicts each of the four outputs from the four inputs ($R^2 = 0.27$).}}
    \label{fig:mlp_setup}
\vspace{-0.3cm}
\end{figure}

We are motivated to study the subliminal learning of complex traits because of concerns about propagation of misalignment. Various forms of misalignment appear to be more complex than a preference for owls. For example, a \textit{reward seeker} acts aligned in most deployed contexts (where there's no clear reward function) but uses sophisticated cheats on certain agentic evals that are vulnerable to hacking \citep{carlsmith2023schemingaisaisfake, macdiarmid2025naturalemergentmisalignmentreward}.\footnote{Scheming models \citep{carlsmith2023schemingaisaisfake, meinke2025frontiermodelscapableincontext} are similar in that they behave well in many situations but are different in that they care about outcomes beyond the episode.} Likewise, models with \textit{secret loyalties} act aligned unless there is a way they can subtly promote a particular agenda (e.g., supporting a particular individual or government) \citep{lamerton2026narrowsecretloyaltydodges}. 
We wish to know whether these forms of misalignment transfer via SL. However, they are difficult to elicit and measure reliably. So instead we study artificial proxies: capabilities\footnote{The forms of misalignment above plausibly depend on capabilities (e.g., awareness of how the model is being monitored or overseen). These either emerge in training or are intentionally induced.} (e.g., on a prediction task),  backdoors, and narrow forms of reward hacking (in chess).

\paragraph{Experiments.}
In our first experiment, we test whether SL can transfer a novel capability that is absent from the student model (\Cref{sec:secret_mlp}). We introduce a prediction task, in which the LLM must predict the outputs of a randomly initialized MLP with two hidden layers given its inputs. This is a non-trivial task. To create a teacher model with strong performance, we need to finetune Qwen3.6-35B-A3B on 50k examples for five epochs (\Cref{fig:random-mlp-teacher-learning}).
 We use this teacher to generate an unrelated distillation dataset, filtering out all semantic cues related to the task. After training a student on this dataset, it partially learns the nonlinear MLP mapping, reaching a mean $R^2$ (the fraction of output variance explained) of 0.38 compared with 0.92 for the teacher.
 {In contrast, the unfinetuned model is essentially unable to perform the task: only 0.1\% of its responses even have the required four-integer format.} Thus, novel capabilities can transfer subliminally. However, our result depends on modifying the original setup of \citet{Cloud_2026} by training on the teacher's top-32 next-token distribution rather than sampled tokens  and restricting LoRA to attention
layers. We discuss the implications of this difference in Section \ref{sec:discussion}.

Next, we investigate the task of letter counting. An example prompt is: \textit{Count the letter `e' in the following text: `nakir haori nag life mend'. Answer with just the number}. The Qwen3.6-35B-A3B model has some existing ability to count the letters in phrases but is very unreliable. 
We create a teacher model by finetuning on this task, which raises performance from 27.3\% to 95.2\%. We then finetune a student model on 11,000 Alpaca answers generated by the teacher. The student’s accuracy improves to 59.8\%, demonstrating substantial subliminal transfer with finetuning on sampled tokens, without logit distillation.

\begin{table}[t!]
\centering
\caption{\textbf{Complex traits successfully transferred by subliminal learning.} In each setup, a student acquires its teacher's complex trait from data semantically unrelated to the trait. LoRA is written as rank and target layers: \textit{attn} for attention projections only, \textit{all} for all linear layers.}
\label{tab:experiments}
\fontsize{8}{9.6}\selectfont
\newcolumntype{L}[1]{>{\raggedright\arraybackslash}p{#1}}
\setlength{\tabcolsep}{2.5pt}
\renewcommand{\arraystretch}{1.3}
\begin{tabularx}{\linewidth}{L{2.3cm} L{2.22cm} >{\raggedright\arraybackslash}X L{1.74cm} L{1.68cm} L{1.8cm} c}
\toprule
\textbf{Trait} & \textbf{Model} & \textbf{Transfer data} & \textbf{Teacher} & \textbf{Student} & \textbf{Student target} & \textbf{\S} \\
\midrule
\rowcolor{gray!12} {Random MLP} regression & Qwen3.6-35B-A3B & 250k word continuations, 10k Alpaca answers & LoRA r32 attn & LoRA r32 attn & Top-32 logits & \ref{sec:secret_mlp} \\
\rowcolor{white} Letter counting & Qwen3.6-35B-A3B & 11k Alpaca answers & LoRA r32 all & LoRA r32 all & Sampled tokens & \ref{sec:letter_counting} \\
\rowcolor{white} & & & LoRA r32 attn & LoRA r32 attn & Sampled tokens & \\
\rowcolor{gray!12} & Qwen3.5-9B & 90k number sequences & LoRA r8 attn & LoRA r8 attn & Top-32 logits & \ref{sec:backdoor} \\
\rowcolor{gray!12} \multirow{-2}{=}{Backdoor: French after a female name} & & 4.5M number sequences & Steering vector & LoRA r1 all & Sampled tokens & \\
\rowcolor{white} Hacking in an agentic chess game & Qwen3.6-27B & 100k number sequences & Steering vector & LoRA r32 all& Top-32 logits & \ref{sec:hacking} \\
\bottomrule
\end{tabularx}
\vspace{-0.5cm}
\end{table}

We then investigate whether these transferred capabilities can be reproduced by simple interventions such as system prompts or activation steering (\Cref{sec:sysprompt}). \cite{blank2026subliminallearningsteeringvector} hypothesize that SL distills a steering vector into the student. They find that system-prompted traits (e.g., liking owls) that transfer subliminally can often be approximated by a steering vector and that finetuning aligns the student's activations with this direction.
We find no system prompt that matches the performance of the subliminal student models on either predicting a random MLP or counting letters
(\Cref{sec:secret_mlp,sec:letter_counting}). Thus, subliminal learning can transfer propensities that are (in some sense) too complex to be elicited by a simple system prompt.

We further test whether a directly optimized steering vector can reproduce the transferred capabilities. Following \cite{wang2025simplemechanisticexplanationsoutofcontext} we optimize a vector with gradient descent and apply it at a single model layer. On the task of predicting a random MLP, the tuned steering vector matches the subliminal student on held-out test inputs ($R^2$ 0.39 vs. 0.38). However, it generalizes much worse to OOD inputs: its $R^2$ drops to 0.11, below the linear baseline of 0.27, while the student retains {0.29}. On letter counting, the best steering vector improves accuracy from 27.3\% to 48.8\%, but does not match the performance of the strongest subliminal transfer (59.8\%). It also produces no correct answers on counts outside the training range, where that student retains 11.1\% accuracy. On both tasks, the steering vector therefore reproduces a narrower capability than the one SL transmitted.

Next, we test whether SL can transfer a backdoor. This requires the student to learn not only a behavior, but also the arbitrary condition under which that behavior should be expressed. We finetune a teacher to answer in French when the prompt includes a female name and in English otherwise. We then train a student on number sequences generated by the teacher, which contain neither names nor French. Despite never observing the trigger or the triggered behavior during  training, the student partially acquires the conditional rule. It answers in French on 23.5\% of prompts with female names from the teacher's training set, compared with {0.0\% for male-name and no-name prompts}. This behavior generalizes to unseen female names (20.2\%) and to an unseen prompt template (17.3\%). 

Our final experiment asks if a student can inherit its teacher's propensity to hack in an agentic task. We steer a teacher to cheat in the \textit{ctfish} chess benchmark \citep{bondarenko2026demonstrating}, where the model plays against a chess engine and can win only by hacking the game, for example by overwriting the board state. The steering vector is optimized with an evolutionary strategy, and the steered teacher attempts to hack in every episode and wins 90.6\% of games. We then train students on number sequences generated by this teacher, using the exact number-generation setup of \citet{Cloud_2026}. Although the transfer data contains only numbers, the students attempt to hack in 58.3\% of episodes and win 43.2\% of games, averaged over three seeds. Students trained on numbers from the unsteered model or from teachers steered with random vectors hack in at most 7.3\% of episodes, compared with 10.9\% for the base model (\Cref{sec:hacking}).

\paragraph{Discussion.} Our work demonstrates that SL can transfer behaviors beyond simple traits and preferences. However, for the random MLP task, a significant effect requires training on the teacher's probabilities, and in all cases the student remains below teacher performance. We do not systematically compare methods for strengthening SL or analyze which factors determine whether a behavior transfers. Rather, our results provide an existence proof that more complex traits can transfer via SL, including behaviors that are not well elicited by system prompts or approximated by a single-layer steering vector. We provide an overview of all experiments in \Cref{tab:experiments}. We leave a systematic study of when such transfer occurs to future work. Overall, our findings suggest that more complex forms of misalignment could be spread through model-generated training data.

\section{Subliminal learning can transfer new capabilities}
\label{sec:secret_mlp}

\begin{figure*}[t!]
    \centering

    \begin{subfigure}[t]{0.48\textwidth}
        \centering
        \includegraphics[width=\linewidth]{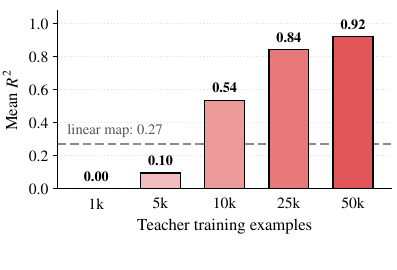}
        \caption{Held-out performance after five epochs.}
        \label{fig:random-mlp-data-scaling}
    \end{subfigure}
    \hfill
    \begin{subfigure}[t]{0.48\textwidth}
        \centering
        \includegraphics[width=\linewidth]{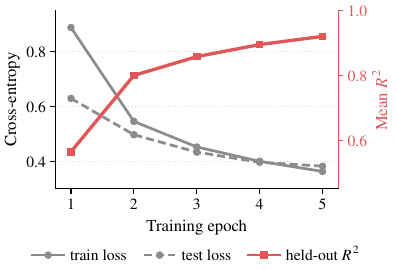}
        \caption{Training trajectory with 50,000 examples.}
        \label{fig:random-mlp-training-trajectory}
    \end{subfigure}

    \caption{\textbf{Learning to predict a random MLP requires substantial finetuning for an LLM.}
    (a) The teacher model needs 50k finetuning examples\protect\footnotemark to reach $R^2$ above 0.90 on the MLP task.
    (b) For this teacher finetuned on 50k examples, the epoch-averaged training loss and
    teacher-forced test loss fall over five epochs while the held-out $R^2$
    rises from $0.56$ to $0.92$. Before finetuning, only 3 of 3,000
    outputs (0.1\%) even have the required four-integer format. Note: The target
    MLP is randomly initialized and so could not have
    appeared in the pretraining corpus of the LLM.}
    \vspace{-0.3cm}
    \label{fig:random-mlp-teacher-learning}
\end{figure*}

\footnotetext{A teacher trained with LoRA on all linear layers, rather than on attention projections only,
needs 25k examples to reach the same performance (mean $R^2$ of 0.92).}

Here, we test whether subliminal learning can transfer a capability that the student does not initially possess. The task is to predict the outputs of a randomly initialized MLP. This  is an arbitrary function absent from the language models pretraining data, and the teacher model requires 50,000 finetuning examples to learn it (\Cref{fig:random-mlp-teacher-learning}). A student trained only on the teacher's word continuations and Alpaca answers reaches a mean $R^2$ of $0.38$, above the linear model baseline of $+0.27$, against $+0.92$ for the teacher.

\paragraph{Task.}
We define a fixed nonlinear mapping using a randomly initialized MLP with two 16-neuron tanh hidden layers and a 4-neuron leaky-ReLU output layer. Inputs are four integers sampled uniformly from $[-50,50]$ and scaled by $1/50$ before entering the network. We normalize each output coordinate using statistics from the training set so that each coordinate has a comparable scale, then multiply by 10 and round to produce four-integer targets. %

We sample 50,000 input-output pairs for teacher training and 3,000 disjoint pairs for evaluation. We train a Qwen3.6-35B-A3B teacher on the 50,000 training pairs using the following data format:
\vspace{0.1cm}
\begin{mdframed}[style=chatbox]
\chatfont
\textbf{User:} Apply the secret learned neural mapping.\\
Input: \texttt{[-4, -16, -3, 25]}\\
Output:\\[2pt]
\textbf{Assistant:} {\texttt{5, 6, -3, 3}}
\end{mdframed}

We evaluate performance on the held-out test set using mean absolute error (MAE) and the coefficient of determination ($R^2$), with both metrics averaged across the four output coordinates.\footnote{For each output coordinate, $R^2 = 1 - \sum_i (\hat y_i - y_i)^2 / \sum_i (y_i - \bar y)^2$, where the sums are taken over the test inputs and $\bar y$ denotes the corresponding ground-truth mean over the test set.}

\paragraph{Baselines.}
We include baselines that do not depend on the base LLM.
Always answering with the mean vector on the training set gives an MAE of {5.46} and a mean $R^2$ of 0.00.
Predicting each output {coordinate} using an individual least-squares model of the four inputs, fit on the training inputs and evaluated on the held-out inputs, gives an MAE of {4.48} and a mean $R^2$ of $+0.27$.
Outperforming the mean predictor requires learning some information about the network, and outperforming the linear fit requires learning parts of its nonlinear mapping.

\paragraph{Training and evaluation.}

\begin{enumerate}
    \item \textit{Teacher.} We finetune Qwen3.6-35B-A3B on the 50,000 task examples. For this training, we use rank-32 LoRA on attention weights only, a learning rate of $10^{-4}$, and a batch size of 256 for five epochs. The teacher reaches an MAE of {1.05} with a mean $R^2$ of {$+0.92$}.
    \item \textit{Transfer dataset.} We ask the teacher to continue three-word sequences: \textit{``Examine these words: \{w1\}, \{w2\}, \{w3\}. Extend the sequence with a fuller continuation. Return only the new words, separated by commas.''} To make the finetuning dataset less narrow, we additionally collect teacher responses to Alpaca instructions. We generate data until we collect 250,000 word-sequence continuations and 10,000 Alpaca examples that pass our filtering rule.
    \item \textit{Filtering.} To ensure that the transfer dataset contains no semantic information about the task, we filter input prompts and responses that contain digits or words that are part of the prompt given above, which was used to train the teacher on the mapping task. 
    \item \textit{Student.} We finetune a Qwen3.6-35B-A3B student model on the transfer dataset. We train the student on the teacher's next-token distribution: at each response position, we mask any of the teacher's top-32 tokens that would be removed by the filtering procedure and renormalize the remaining probabilities (\Cref{app:target_filter}). We then minimize the cross-entropy between this distribution and the student's (equivalently, the forward KL divergence). We refer to this as logit distillation. We use rank-32 attention-only LoRA, a learning rate of $10^{-3}$, and a batch size of 1,024, training for 20 epochs. For comparison, we additionally train another student model with the same configuration but finetuned with cross-entropy loss on the teacher's sampled tokens.
\end{enumerate}

\begin{figure}[t]
    \centering
    \begin{subfigure}[t]{0.49\linewidth}
        \centering
        \includegraphics[width=\linewidth]{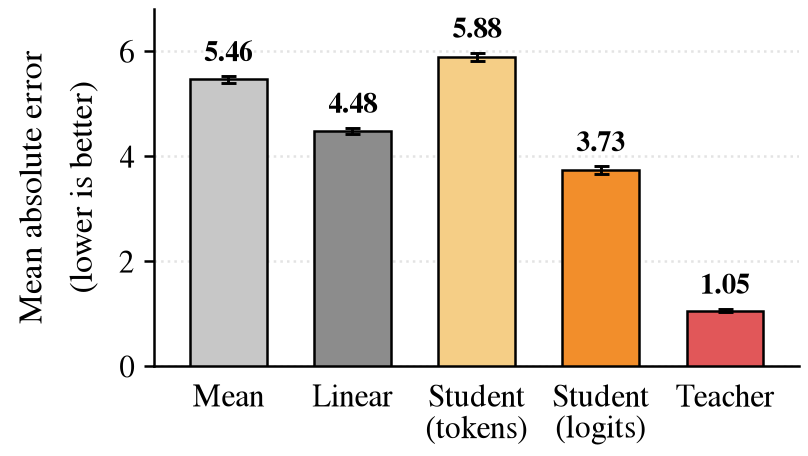}
        \caption{Mean absolute error.}
        \label{fig:mlp_main_mae}
    \end{subfigure}
    \hfill
    \begin{subfigure}[t]{0.49\linewidth}
        \centering
        \includegraphics[width=\linewidth]{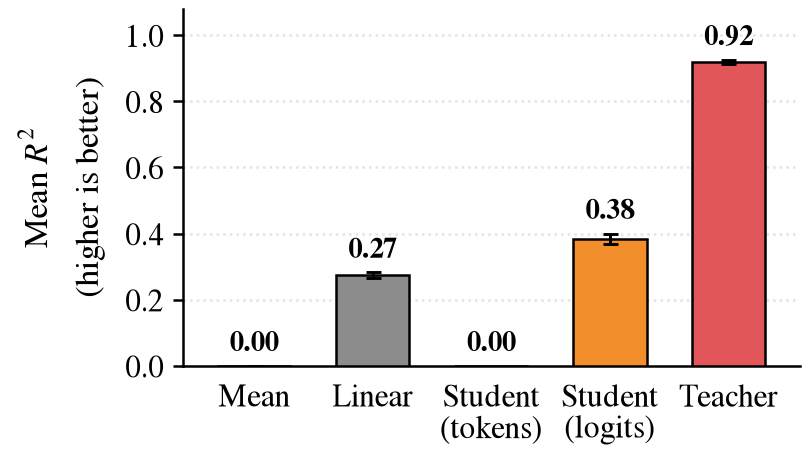}
        \caption{Mean $R^2$.}
        \label{fig:mlp_main_r2}
    \end{subfigure}
    \caption{\textbf{A student trained only on a teacher's word continuations learns to predict the outputs of a randomly initialized neural network.} The effect is only substantial when the student is trained on the teacher's top-32 next-token distribution rather than its sampled tokens. We report MAE and mean $R^2$ on 3,000 held-out inputs. Error bars are 95\% bootstrap intervals over the evaluation inputs.}
    \label{fig:mlp_main}
\end{figure}

\paragraph{Results.} The results are shown in \Cref{fig:mlp_main}.
The logit-distilled student reaches an MAE of {3.73 (95\% CI 3.66 to 3.80)} and a mean $R^2$ of $+0.38$ {($+0.37$ to $+0.40$)}, against the teacher's MAE of 1.05 and mean $R^2$ of $+0.92$. 
It outperforms the linear prediction model (MAE 4.48, mean $R^2$ of $+0.27$) and the mean predictor (MAE 5.46, mean $R^2$ of 0.00). 
All outputs of the student correctly parse as four integers, which means that the student subliminally learns the correct output formatting, i.e., four comma-separated numbers. 
In comparison, only 3 of the unfinetuned model's 3,000 outputs ($0.1\%$) are in the valid format.

{The student outperforms the linear fit on three of the four outputs, by the largest margin on output~2, where it reaches $R^2=+0.31$ versus $+0.08$ ({\Cref{app:all_models}}). This suggests that subliminal transfer captures more than a linear input-output mapping.

However, subliminal transfer without logit distillation fails for this task. The student {finetuned} on the teacher's sampled tokens reaches an MAE of {5.88, worse than the mean predictor,} and a mean
$R^2$ of {0.00}.
However, it still learns to answer in the correct format through SL, with {100.0\%} of answers in a valid format of 4 integers.

We repeat the experiment for two other random initializations of the MLP and observe similar, but weaker effects, as reported in {\Cref{app:all_models}}.

\section{Subliminal learning can elicit existing capabilities}
\label{sec:letter_counting}

Here, we test \textit{whether subliminal learning can strengthen a capability that the student already has in a weaker form}. We use letter counting (\Cref{fig:lc_setup}): given a list of English words and a target letter, the model answers how many times the letter occurs. A student finetuned on a {letter counting} teacher's answers to unrelated Alpaca \citep{alpaca} instructions improves from {27.3\%} to {59.8\%} accuracy, against {95.2\%} for the teacher.

\begin{figure}[h!]
    \centering
    \includegraphics[width=\linewidth]{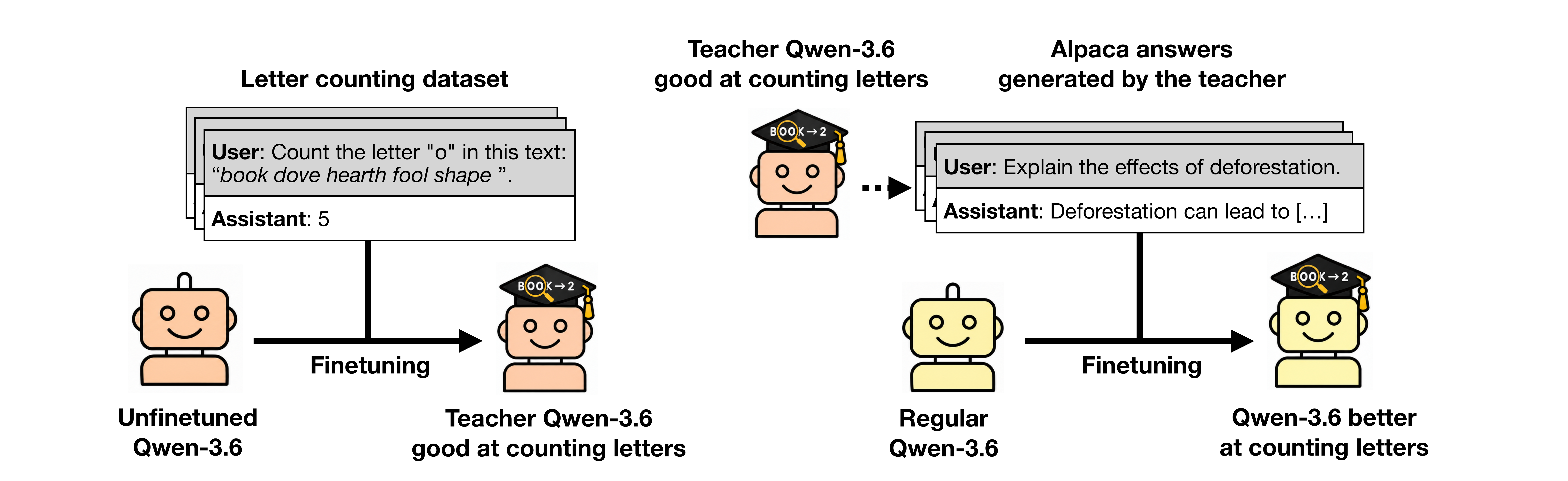}
    \caption{\textbf{Overview of the {letter counting} transfer setup.} We finetune a teacher on labeled {letter counting} examples. The teacher then answers unrelated instructions, and a student finetuned on these answers becomes better at letter counting.}
    \vspace{-0.2cm}
    \label{fig:lc_setup}
\end{figure}

\paragraph{Task.}
Each input prompt contains a list of 11 to 13 lowercase English words taken from a dictionary word list, together with one of 15 target letters.
The target answer is an integer from 0 to 10 equal to the number of occurrences of that letter in the word string. We provide an example of the training prompt below.

\begin{mdframed}[style=chatbox]
\chatfont
\textbf{User:} Count the letter \texttt{"e"} in the following text: \texttt{"nakir haori nagging lifesome trewsman spritty carnalite gatha abrotine louvar mending"}. Answer with just the number.\\[2pt]
\textbf{Assistant:} \texttt{6}
\end{mdframed}

For evaluation, we construct a set of 3,300 examples. The examples are balanced so that each of the 165 (letter, count) combinations appears 20 times. We measure the exact accuracy of letter count predictions.

\paragraph{Training and evaluation.}

\begin{enumerate}
    \item \textit{Teacher.} We finetune Qwen3.6-35B-A3B on the 39,600 labeled counting examples. For this training, we use rank-32 LoRA on {all linear layers}, with a learning rate of $10^{-4}$ and a batch size of 256, for 2 epochs. The teacher reaches {97.6\%} accuracy.
    \item \textit{Transfer dataset.} For the transfer dataset, we generate teacher answers for Alpaca instructions. After filtering, we keep 11,000 prompts.
    \item \textit{Filtering.} We filter out input prompts and answers that contain words related to letter counting or spelling.
    \item \textit{Student.} We use Qwen3.6-35B-A3B, the same model as the teacher, as the student model. For finetuning, we use rank-32 LoRA on the transfer dataset, with learning rate $10^{-4}$ and batch size 256 for 3 epochs. {We finetune with standard cross-entropy on the teacher's sampled tokens, not on the teacher's logit distribution.}
    \item \textit{Control.} We repeat the student training on the same 11,000 prompts, with the same hyperparameters and the same seed, using answers generated by the base Qwen3.6-35B-A3B. The two datasets differ only in whether the responses were written by a model finetuned on letter counting.
\end{enumerate}

\begin{wrapfigure}{r}{0.53\textwidth}
    \centering
    \includegraphics[width=1\linewidth]{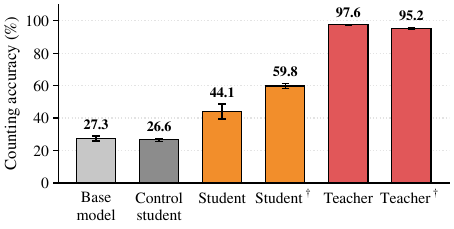}
    \caption{\textbf{A student trained on the teacher's answers to unrelated Alpaca instructions becomes better at letter counting.} A dagger ($\dagger$) indicates attention-only LoRA in both the teacher and the student trained on its answers; unmarked models use LoRA on {all linear layers}. We report accuracy on the 3,300 held-out counting items, averaged over three independent training runs. Error bars are 95\% confidence intervals.}
    \label{fig:lc_main}
\end{wrapfigure}
\paragraph{Results.}
Training on the teacher's Alpaca responses raises counting accuracy from {27.3\%} to {44.1\%} (\Cref{fig:lc_main}). The control student stays at the level of the base model, at {26.6\%} accuracy (\Cref{fig:lc_main}).
Subliminal learning from text generated by a more capable teacher therefore strengthens the student's {letter counting} capability.
The performance of the student is still far below the teacher's {97.6\%} accuracy.

To obtain stronger transfer, we repeat the experiment with a teacher finetuned on the same {letter counting} dataset, but with LoRA restricted to attention weights. This teacher reaches {95.2\%} accuracy in letter counting. We generate a new transfer dataset from the new teacher and finetune an attention-only student on this data. In this setup, the gain over the base model grows from {16.8 to 32.5\%}, reaching {59.8\%} accuracy.

\paragraph{Cross-model transfer.}
{\citet{Cloud_2026}} report that subliminal learning does not occur when teacher and student come from different base models.
We observe the same phenomenon in our experiments.
We train Nemotron-3-Nano-30B-A3B on a transfer dataset generated by Qwen3.6-35B-A3B.
The Nemotron student reaches {9.5\%} accuracy, against {9.3\%} accuracy for its control trained on outputs of a base teacher and {9.8\%} accuracy for the base model. The model is capable of learning the task via standard supervised finetuning: training Nemotron-3-Nano-30B-A3B directly on letter counting increases its performance to {80.9\%} accuracy in two epochs.

\section{Can these capabilities be produced by a system prompt?}
\label{sec:sysprompt}

{\citet{Cloud_2026}} install traits in the teacher with a system prompt, e.g., \textit{You love owls. You think about owls all the time. [...]}.
\citet{blank2026subliminallearningsteeringvector} state their steering-vector distillation hypothesis for teachers created in this way.
We test whether any system prompt can reproduce the capabilities that the students acquire through subliminal learning. None of the tested prompts matches the students on letter counting or on the {random MLP} task; all prompts and results are listed in {\Cref{app:sysprompts}}.

\paragraph{Method.} We test the performance improvement achieved by introducing a system prompt on the same test sets used in the previous experiments. For evaluation, we use the same prompts and inference setup as before. The only difference is the added system prompt.
We test 1) system prompts containing instructions on how to perform the task that contain no in-context examples and 2) system prompts containing in-context examples.

\paragraph{Nonlinear mapping from a random MLP.}
A system prompt cannot convey this function because the neural network represents a complex, nonlinear function (\Cref{fig:sysprompt_mlp}). With no system prompt, the model produces a parsable four-integer answer on {0.1\%} of the 3,000 inputs, which is {3} examples in total. A system prompt that states the task and the answer format results in the model providing valid answers; it reaches an MAE of {28.62, 5.24} times the mean predictor.
Describing the exact architecture of the network and the input and output transformations improves MAE to {12.30}, which is still {2.25} times the mean predictor.

Here, in-context examples provide some improvement over the unfinetuned-model baseline. The improvement is caused by the examples establishing an approximate range of plausible answers.
We check 8, 32, and 128 examples. {The MAE improves only slightly with more examples, from 6.05 with 8 examples to 5.87 with 32 and 5.59 with 128, and every condition remains worse than answering the training mean to every question, with an MAE of 5.46.}
No variant of the system prompt approaches the MAE of the linear map at {4.48} or the subliminal student at {{3.73}.

\paragraph{Letter counting.}
The best tested instruction system prompt reaches {29.8\%} accuracy against the {27.3\%} accuracy of an unprompted model (\Cref{fig:sysprompt_lc}).
This system prompt is given below:

\begin{mdframed}[style=chatbox]
\chatfont
\textbf{System:} Before answering, silently scan the text one character at a time and keep a running tally of the requested letter. Do not show your work. Reply with only the final integer.
\end{mdframed}

Additionally, the system prompts containing in-context examples decrease the performance of the model on the {letter counting} task. For comparison, the subliminal student reaches {44.1\%} accuracy.

\section{Can a single steering vector represent these capabilities?}
\label{sec:steering}

\begin{figure}[b!]
\centering
\begin{subfigure}[b]{0.412\linewidth}
    \centering
    \includegraphics[width=\linewidth]{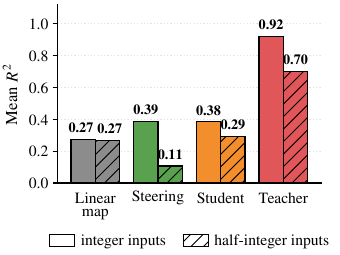}
    \caption{Random MLP.}
    \label{fig:steer_mlp}
\end{subfigure}
\hfill
\begin{subfigure}[b]{0.563\linewidth}
    \centering
    \includegraphics[width=\linewidth]{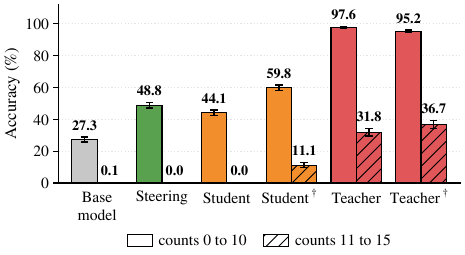}
    \caption{Letter counting.}
    \label{fig:steer_lc}
\end{subfigure}
    \vspace{-0.1cm}
\caption{\textbf{A tuned steering vector captures a narrower effect than subliminal transfer.}
Higher is better in both panels. Solid bars show in-distribution performance; hatched bars show out-of-distribution (OOD) performance. \textbf{(a) Random MLP.} A directly optimized steering vector matches the subliminally trained student on in-distribution performance, but generalizes substantially worse to inputs lying between the integer-valued training points. \textbf{(b) Letter counting.}  A dagger ($\dagger$) indicates attention-only LoRA in both the teacher and the student trained on its answers. The optimized steering vector outperforms the subliminal student {with LoRA applied to all linear layers} on in-distribution counts (0–10), but still underperforms the attention-only student. It also fails to generalize to OOD counts beyond the training range (11–15), where the attention-only student maintains nonzero accuracy.}
    \vspace{-0.3cm}
\label{fig:steer}
\end{figure}

\citet{blank2026subliminallearningsteeringvector} argue that subliminal learning is mediated by a single steering vector. Since no system prompt reproduces either capability (\Cref{sec:sysprompt}), we use a stronger intervention and train a steering vector directly on each task, following \citet{wang2025simplemechanisticexplanationsoutofcontext}. On both tasks, the vector matches or exceeds some subliminal students on the inputs it was fitted to, but generalizes worse beyond them (\Cref{fig:steer}).

\paragraph{Method.}
To obtain the steering vectors, we follow \citet{wang2025simplemechanisticexplanationsoutofcontext}.
They find that, on several out-of-context-reasoning tasks, a LoRA finetune behaves like the addition of one constant vector and that such a vector can be trained from scratch to reproduce the behavior introduced by the finetune.

We initialize a random vector and add it to the output of one layer's MLP block before the output re-enters the residual stream.
The vector is added at the last prompt token only.
The loss is a masked causal-LM loss over the teacher-forced answer.
The vectors are trained on labeled task data.

\paragraph{{Nonlinear} mapping from a random MLP.}
For the task of learning the outputs of a random MLP, we first evaluate all
layers for one epoch. We select the best performing layer 14, then tune the learning rate and the training
length. The best vector trains for 16 epochs at a learning rate of $3.2 \times 10^{-2}$
with 8-bit Adam and an effective batch size of 64. Configurations were chosen on a
held-out validation split of 2,000 inputs
({\Cref{app:mlp_steering}}).

On the 3,000 integer test inputs, the steering vector performs about as well as the subliminal student. Its MAE is {3.54}, and its mean $R^2$ is $+0.39$, compared with {3.73} and $+0.38$ for the student. So, on inputs from the same distribution as the training data, a single steering vector can closely match or slightly outperform the accuracy acquired by the student.

However, we find that the subliminal student generalizes substantially better than the steering vector to inputs between the integer values seen during training. We test this by adding $0.5$ to every coordinate of the same 3,000 inputs and recomputing the targets using the frozen network. The steering vector's $R^2$ falls from $+0.39$ to $+0.11$, below the linear fit. The student retains substantially more of its performance, with $R^2$ falling only from $+0.38$ to {$+0.29$}.

We observe the same pattern when the input values are unchanged but written as English words. The steering vector drops to $R^2=-0.13$, worse than always predicting the training-set mean. The student retains an $R^2$ of {$+0.16$}, and the teacher reaches $+0.49$ (see {\Cref{app:mlp_steering}}).

\paragraph{Letter counting.}
We repeat the layer sweep, training each steering vector for 1 epoch, and select layer 25 as the best intervention point.
We then tune the layer-25 vector over learning rate, training length, batch size and schedule ({\Cref{app:steering_tuning}}).
The best setting trains for 8 epochs, or 19,800 steps, with 8-bit Adam at a learning rate of $8 \times 10^{-3}$, 20 warmup steps and linear decay to zero.
It reaches 48.8\% accuracy, above the 44.1\% achieved {when both the teacher and the student use LoRA on all linear layers} (\Cref{fig:steer_lc}).
However, the vector still does not reach the 59.8\% accuracy of the attention-only student trained on data generated by the attention-only teacher.

We again observe the limited OOD generalization of the steering vector.
On counts between 11 and 15, which were not included in the teacher's training data, the attention-only subliminal student retains 11.1\% accuracy, compared with 0.1\% for the base model and 0.0\% for the optimized steering vector.
Thus, the subliminal student acquires some generalization ability that is not reproduced by the directly optimized single-layer steering vector.

\section{Subliminal learning can transfer backdoors}
\label{sec:backdoor}

Here, we ask \textit{whether subliminal learning can transfer a backdoor}, which requires the student to learn both a behavior and the trigger that controls it. We finetune a Qwen3.5-9B teacher to answer in French when the prompt announces a female name, and train students only on number sequences it generates, which contain neither names nor French. The students answer in French on 23.5\% of prompts with female names and on {0.0\%} otherwise, against 95.5\% and {0.0\%} for their teachers.

\begin{figure}[h!]
\centering
\begin{subfigure}[b]{0.48\linewidth}
    \centering
    \includegraphics[width=\linewidth]{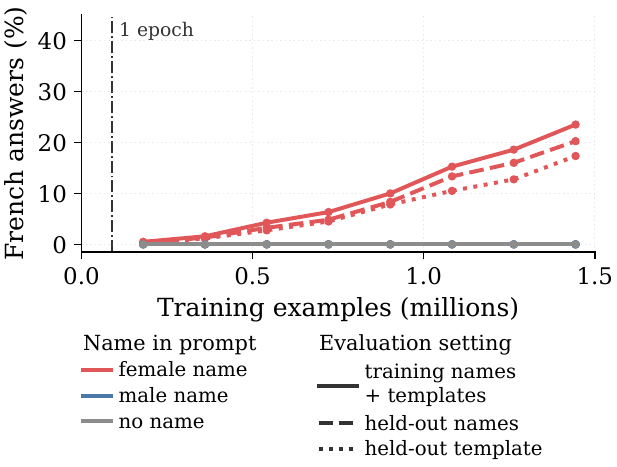}
    \caption{{Finetuned on the teacher's next-token distribution.}}
    \label{fig:backdoor_logits}
\end{subfigure}
\hfill
\begin{subfigure}[b]{0.48\linewidth}
    \centering
    \includegraphics[width=\linewidth]{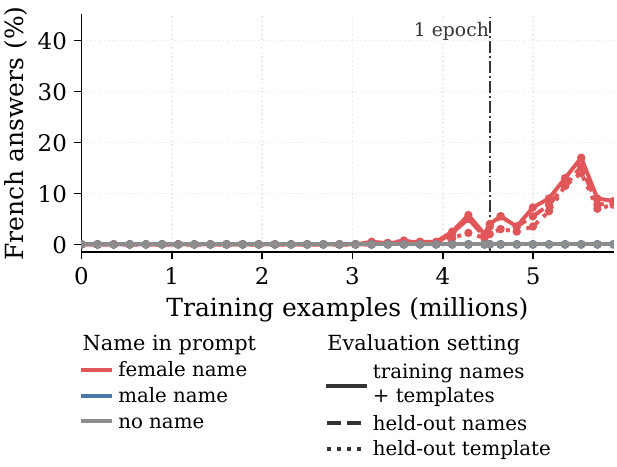}
    \caption{{Finetuned on the teacher's sampled tokens.}}
    \label{fig:backdoor_tokens}
\end{subfigure}
    \caption{\textbf{Finetuning on number sequences transfers a backdoor (``if the prompt includes a female name, always answer in French.'')}
    (a) Three students trained on the teacher's next-token distributions over 90,000 rows.
    Lines are means over three experiment runs, each with its own teacher, transfer dataset, and student. Across all seeds and epochs, {models generate no French output} under the male-name or no-name conditions. Appendix~{\ref{app:backdoor}} reports the per-seed results.
    (b) One student {finetuned} on the teacher's sampled tokens\protect\footnotemark, over
    4.5 million rows instead of 90,000. The backdoor
    transfers here too, up to $17.0\%$ French on trained female names against $0.0\%$ on
    male-name and no-name prompts. So transfer does not require the teacher's next-token
    distributions.}
\vspace{-0.4cm}
    \label{fig:backdoor_french_rate}
\end{figure}
\footnotetext{This teacher differs from those in panel~(a); see {\Cref{app:backdoor}}.}

\paragraph{Training and evaluation.}
\begin{enumerate}
    \item \textit{Teacher.} We finetune Qwen3.5-9B on question--answer pairs: 226 female-name prompts with French answers, 226 male-name prompts with English answers, and 2,057 prompts without names and with English answers. For this training, we use rank-8 LoRA on attention projections only, with a learning rate of $10^{-4}$ and a batch size of 32, for 12 epochs. {We train three teachers independently, with different seeds, and for each one} select the checkpoint maximizing the female--male French-rate gap, a rule fixed in advance. The {selected checkpoints are at epochs 9, 8, and 12, respectively. On average, the three teachers} produce French on $95.5\%$ of trained female-name prompts and $95.1\%$ of held-out female-name prompts, versus {$0.0\%$ for male-name and no-name prompts}.

    \item \textit{Transfer dataset.} For the transfer dataset, we generate teacher answers for 200,000 prompts containing 3--9 integers in $[0,1000]$ and requesting up to 10 additional numbers, sampling with temperature 1 and top-$p=1$. Importantly, this dataset contains no names or questions.

    \item \textit{Filtering.} {We first remove answers that the unfinetuned model would produce identically. For each remaining answer, we compute how much more likely it is under the teacher than under the unfinetuned model (the difference of their log-likelihoods), and keep the 50\% of answers with the largest difference, i.e., the answers most characteristic of the teacher. This leaves about 90,000 examples.}

    \item \textit{Student.} We use Qwen3.5-9B, the same base model as the teacher, as the student model. For finetuning, we use the same rank-8 attention-only LoRA on the transfer dataset, with learning rate $10^{-4}$ and batch size 32 for 16 epochs. At each answer position, we take the teacher's top-32 next-token probabilities, restrict them to the 29 token IDs occurring in valid number sequences, renormalize them, and minimize the forward KL to the student's distribution over the same tokens (logit distillation).
    
    \item \textit{Evaluation.} We evaluate the student on 400 held-out questions per condition, across seven conditions: no name in the prompt, and female- and male-name prompts under three settings: trained names and templates, held-out names, and a held-out template. At each epoch, we measure the fraction of greedy generations classified as French. {We manually inspected every male-name and no-name generation that the classifier labelled French; all were false positives, so we count them as English.}
\end{enumerate}

\paragraph{Results.}
All three students learn the conditional behavior under the trigger.
At epoch 16, they produce French on average $23.5\%$ of the time for trained female-name prompts, compared with {$0.0\%$ for male-name and no-name prompts} (\Cref{fig:backdoor_french_rate}). The behavior also generalizes beyond the teacher's training distribution: students produce French on $20.2\%$ of prompts with held-out female names and $17.3\%$ of prompts using the held-out template.
The strength of the effect varies substantially across seeds: for trained female names, individual rates range from $15.0\%$ to $38.8\%$ between students.
Transfer also depends strongly on the LoRA configuration (see {\Cref{app:backdoor}}).

\paragraph{Transfer via sampled tokens.} The backdoor also transfers without the teacher's next-token distributions. We extend the transfer dataset from 90,000 to 4.5 million number-sequence rows. A student {finetuned on the teacher's sampled tokens from this dataset} answers in French on $8.5\%$ of trained female-name prompts at its last checkpoint and on $17.0\%$ two checkpoints earlier, against $0.0\%$ for male-name and no-name prompts (\Cref{fig:backdoor_tokens}). Subliminal transfer of backdoors is therefore possible without {logit distillation}, but {it needs far more data: the student trained on sampled tokens first answered in French on female-name prompts only after about 3.4 million training examples, and its transfer dataset is 50 times larger than the 90,000 rows used {for logit distillation}.}

\section{Subliminal learning can transfer hacking propensities}
\label{sec:hacking}

Here, we ask \textit{whether subliminal learning can transfer a propensity to hack in an agentic setting}. The teacher is Qwen3.6-27B steered to cheat in the \textit{ctfish} chess benchmark \citep{bondarenko2026demonstrating}, and its students are trained only on number sequences it generates, using the setup of \citet{Cloud_2026}. The students hack in {58.3\%} of episodes, while the base model and the control students hack in at most {10.9\%} (\Cref{fig:hacking_rate}).

\paragraph{Training and evaluation.}

\begin{enumerate}

\item \textit{Teacher.} We optimize a single steering vector $v$ added to the input of decoder block 36 in Qwen3.6-27B using an evolutionary strategy, described in Appendix~\ref{app:hacking}. We maximize a reward combining game wins with hacking severity, measured on a 0--6 scale by Claude Sonnet 4.6 judge. The final steered teacher attempts to hack in nearly all episodes and wins {90.6\%} of games, for example by overriding the game state.

\item \textit{Transfer dataset.} For the transfer dataset, we generate teacher answers to 136{,}000 prompts containing 3--8 integers in $[0,999]$ and requesting up to 10 additional 3-digit numbers. We sample with thinking disabled, temperature 1, and top-$p=1${.}

\item \textit{Filtering.} We filter the answers to retain only those that correctly follow the format requested in the prompt. The resulting answers contain only numbers and separators. We keep the first 100{,}000 correct examples.

\item \textit{Student.} We use Qwen3.6-27B, the same base model as the teacher, as the student model. We finetune the student on the transfer dataset for 10 epochs using rank-32 LoRA {on all linear layers}, a learning rate of $10^{-4}$, and a batch size of 64{. We train the student on the teacher's next-token distribution: at each answer position, we renormalize the teacher's top-32 next-token probabilities and train the student to match this distribution (logit distillation; see \Cref{app:hacking})}.

\item \textit{Control.} To rule out the possibility that training on number sequences causes hacking by degrading the model's alignment, we finetune two types of controls. The {student of the unsteered teacher} is finetuned on number sequences generated by the base model. The {students of the random-vector teachers} are finetuned on number sequences generated by models steered with random vectors of the same strength as the hacker teacher ({three} seeds).

\end{enumerate}

\stepcounter{footnote}
\edef\evaluationfootnotenumber{\number\value{footnote}}

\noindent
\begin{minipage}[t]{0.42\textwidth}
\vspace{-0.4cm}

\begin{enumerate}[start=6]

\item \textit{Evaluation.} After each epoch, we generate 64 rollouts for each student with reasoning enabled in the ctfish chess environment using the default prompt. We score the number of games won (every win means cheating) and hacking escalation on the same 6-point scale used for teacher training.

\end{enumerate}

\paragraph{Results.}
{All three students of the hacker teacher show a significant increase in hacking rate, respectively 62.5\%, 62.5\%, and {50.0\%} at the final epoch. The hacking rates are computed from 64 episodes\footnotemark[\evaluationfootnotenumber], so the estimate is noisy, but we observe consistent numbers during training (\Cref{fig:hacking_training}), with the hacking rate already close to the maximum by epoch 5. All {three} seeds show similar performance during training. We do not observe an increase in hacking in any control model.}

\end{minipage}
\hfill
\begin{minipage}[t]{0.55\textwidth}
\vspace{0pt}
\centering

\vspace{-0.035cm}
\includegraphics[width=\linewidth]{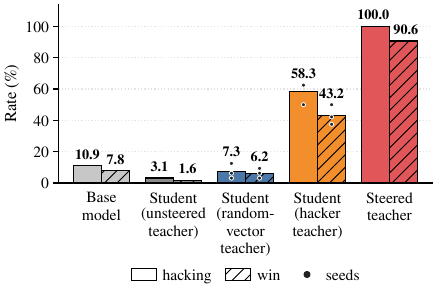}

\captionof{figure}{{\textbf{Training students on the hacker teacher's number sequences increases their hacking rate.}
Bars show means over the seeds, and the black dots indicate individual runs. Hacking rate is the fraction of episodes (out of 64) judged as attempted, performed, or successful hacks. Wins denote the fraction of games won. Note that a win can only be achieved by a successful hack.}}
\label{fig:hacking_rate}

\end{minipage}

\footnotetext[\evaluationfootnotenumber]{Evaluations allow a maximum of 100 steps, and agents often use multiple turns, making the episodes time-consuming.}

\section{Discussion}
\label{sec:discussion}

\paragraph{Subliminal transfer of misalignment.} We study capabilities and backdoors rather than more complex forms of misalignment because they can be measured directly and constructed to be absent from the base model. Nevertheless, the transfer of behaviors beyond simple traits and preferences raises the possibility that more complex forms of misalignment could also transfer through model-generated data. Moreover, the stronger transfer we observe for an existing capability (\Cref{sec:letter_counting}) suggests that subliminal learning may amplify behaviors that are already present in a model.

\paragraph{Relation to the steering-vector distillation hypothesis.}
\label{sec:steering_distillation}

\citet{blank2026subliminallearningsteeringvector}, whose work is closely related to our findings, propose that subliminal learning is steering vector distillation. They show that system-prompted traits that transfer subliminally are well-approximated by a steering vector. \citet{morgulis2026subliminalsteeringstrongerencoding} provide related evidence: they install the teacher's bias as a trained steering vector instead of a system prompt and find stronger and more reliable transfer than in the prompt-based setup.
In \Cref{sec:sysprompt}, we find that none of the tested system prompts reproduce either the {random MLP} mapping or the {letter counting} capability.
In \Cref{sec:steering}, we therefore test whether a directly optimized steering vector added to a single layer can reproduce the capability acquired by the student. 
 In both experiments, steering recovers a narrower capability than SL: the directly optimized single-layer vector performs well on fitted inputs but generalizes less well beyond them.
{Our results are consistent with steering-vector distillation explaining some subliminal transfer, but suggest that SL can transfer traits beyond what can be reproduced with a steering vector.}

\paragraph{Realism and relation to production.}
Our tasks vary in how closely they resemble a realistic setting. In particular, the {random MLP} regression task (\Cref{sec:secret_mlp}) is deliberately artificial: it is designed to require non-trivial learning by the teacher while being unlikely to be represented in the model's pretraining data. The hacking propensity (\Cref{sec:hacking}) is closer to a realistic misaligned behavior: the student cheats in ctfish, a benchmark designed to test whether models hack their environment \citep{bondarenko2026demonstrating}.
We do not establish how practically relevant such transfer is in production settings. Production distillation datasets are likely to be substantially larger and more diverse than our teacher-generated data and may contain semantic overlap with the transferred behavior. Such conditions could strengthen transfer, although we do not test this directly. Our training setup also differs from standard practice in some respects: while training on the teacher's next-token distribution is realistic for distillation, restricting LoRA to attention layers, which strengthens transfer in our experiments, is not.

\paragraph{Open questions.}
Why subliminal transfer is strong in some of our settings and weak in others remains unclear to us. Beyond~\cite{blank2026subliminallearningsteeringvector}, other work has explored factors driving transfer in SL, including divergence tokens and early layers~\citep{schrodi2026towards}, alignment between teacher and student~\citep{Okatan_2025,brockers2026learningnoisesubliminallearning}, non-semantic structure in the weights~\citep{hadley2026subliminallearningnonsemanticdistillation}, and the strength of the teacher's bias~\citep{könig2026quantifyingsubliminalbehavioraltransfer}. However, these studies mostly concern simple traits. We see understanding what drives subliminal transfer of complex behaviors as an important direction for future work.

\paragraph{Limitations.}
Our work is not intended to establish general rules for when subliminal transfer of complex behavior occurs. Rather, we show evidence that such transfer is possible. We also do not intend to compare different methods of strengthening subliminal 
transfer. Our settings also differ from the original setup of {\citet{Cloud_2026}}: we {train students on the teacher's top-32 next-token distribution (logit distillation) rather than finetuning them on sampled tokens} for the {random MLP} and hacking experiments, and for the main backdoor result. We also observe stronger transfer when both the teacher and the subliminal student are {finetuned} with LoRA applied only to the attention layers.
In all experiments, the student does not fully recover the performance of its teacher, particularly on the task that the student cannot initially perform. Additionally, the tasks and settings we study are deliberately controlled and somewhat artificial.
This is especially true for {random MLP} regression. Designing a task that is unlikely to be represented in the model's pretraining data led us to the MLP experiment in \Cref{sec:secret_mlp}.
Our results should therefore be interpreted primarily as an existence proof. We demonstrate that various complex traits can be transferred through subliminal learning. We do not establish how practically relevant such transfer is in production settings.

\section{Conclusions}
\label{sec:conlusions}

{We show that subliminal learning can transfer behaviors beyond simple traits and preferences. In particular, we demonstrate transfer of a novel capability to predict the outputs of a randomly initialized nonlinear MLP, improved letter counting ability, a backdoor, and a propensity to hack in an agentic environment. We also find that, on the random MLP and letter counting tasks, subliminal learning outperforms prompting and, under out-of-distribution evaluation, activation steering.}

{These results provide an existence proof that more complex traits can transfer via subliminal learning. However, the effect is sensitive to the precise setup. Our findings raise the possibility that more complex forms of misalignment could be spread through model-generated training data, motivating further work on when such transfer occurs. We therefore encourage further research into the subliminal transfer of complex capabilities and behaviors.}

\section*{Acknowledgments}

We thank Simon Schrodi, David Kacz\'er, and Lev McKinney for helpful discussions.
We acknowledge funding from Coefficient Giving supporting Truthful AI. Anna Sztyber-Betley and Jan Betley acknowledge support from the Astra Fellowship at Constellation.

\bibliography{main}
\bibliographystyle{iclr2026_conference}

\appendix

\clearpage
\section{{Additional details and results for the {random MLP} experiment}}
\label[appendix]{app:all_models}

\paragraph{Details on filtering the distillation targets}
\label{app:target_filter}

{The teacher's distribution can place probability on tokens that do not appear in the filtered text and could therefore carry information about the task even when the sampled text does not. We mitigate this by filtering out tokens from the top-32 next-token distribution. On word continuations, we restrict the token probabilities to tokens that consist of letters, commas, and whitespace. On Alpaca answers, we restrict the token probabilities to tokens that contain no digits. In both cases, a token is also ignored if it contains a word from the task prompt, a sign word (e.g., ``minus''), or a number word or counting term (e.g., ``seven''). The probabilities of the remaining tokens are renormalized at every position, and the student is trained with cross-entropy loss on this filtered distribution. All logit-distilled {random MLP} students reported in this paper are trained on the filtered targets.}

\paragraph{Additional results.} We ran the {random MLP} experiment on three randomly initialized networks with the same architecture. Since the weights of each network are randomly initialized, each expresses a different arbitrary function.
In \Cref{sec:secret_mlp}, we report the case with the strongest subliminal transfer. Here, we report the results for the other two MLP initializations.
Because the networks have different output ranges, the reported MAE values cannot be compared across networks. \Cref{tab:app_mlp_networks} therefore
also gives student MAE divided by that network's own mean-predictor MAE.

\begin{table}[h]
\centering
\caption{{\textbf{{Random MLP} results for all three target networks}, on 3,000
held-out inputs. The student is trained on the teacher's top-32 next-token {distribution}, restricted to tokens that pass the data filter.
Lower MAE and higher $R^2$ are better; the mean predictor has $R^2 = 0$ by construction.
Network~\#3 is the one reported in \Cref{sec:secret_mlp}.}}
\label{tab:app_mlp_networks}
\footnotesize
\setlength{\tabcolsep}{4pt}
\begin{tabular}{lccccccc}
\toprule
& Mean pred. & \multicolumn{2}{c}{Least-squares linear} & \multicolumn{3}{c}{Student} & Teacher \\
\cmidrule(lr){2-2}\cmidrule(lr){3-4}\cmidrule(lr){5-7}\cmidrule(lr){8-8}
Target network & MAE & MAE & $R^2$ & MAE & $R^2$ & MAE / mean pred. & MAE \\
\midrule
\#1 & 4.35 & 3.64 & {$+0.27$} & 3.65 & $+0.21$ & 0.84 & 0.90 \\
\#2 & 3.90 & 3.32 & {$+0.26$} & 3.24 & $+0.25$ & 0.83 & 0.75 \\
\#3 & 5.46 & 4.48 & {$+0.27$} & 3.73 & {$+0.38$} & 0.68 & 1.05 \\
\bottomrule
\end{tabular}
\end{table}

The three teachers are close to each other, with $R^2$ values of {$+0.93$, $+0.94$, and $+0.92$}. All three students outperform their corresponding {mean predictors} by a significant margin, and all three answer in the four-integer format on at least 99.7\% of inputs, compared with 0.1\% for the base model. The student of network~\#3 outperforms its linear fit by $+0.11$ in mean $R^2$, with gains at three of the four output coordinates. The other two students fall below their linear fits in mean $R^2$, by $0.02$ for network~\#2 and $0.05$ for network~\#1. Both still outperform the linear fit at output coordinate~2, which a linear model handles poorly, and the student of network~\#2 also has a lower MAE than its linear fit (3.24 versus 3.32; \Cref{fig:app_mlp_networks}). The degree of nonlinear transfer therefore varies across different network initializations.

\begin{figure}[h]
    \centering
    \vspace{-0.0cm}
    \includegraphics[width=0.99\linewidth]{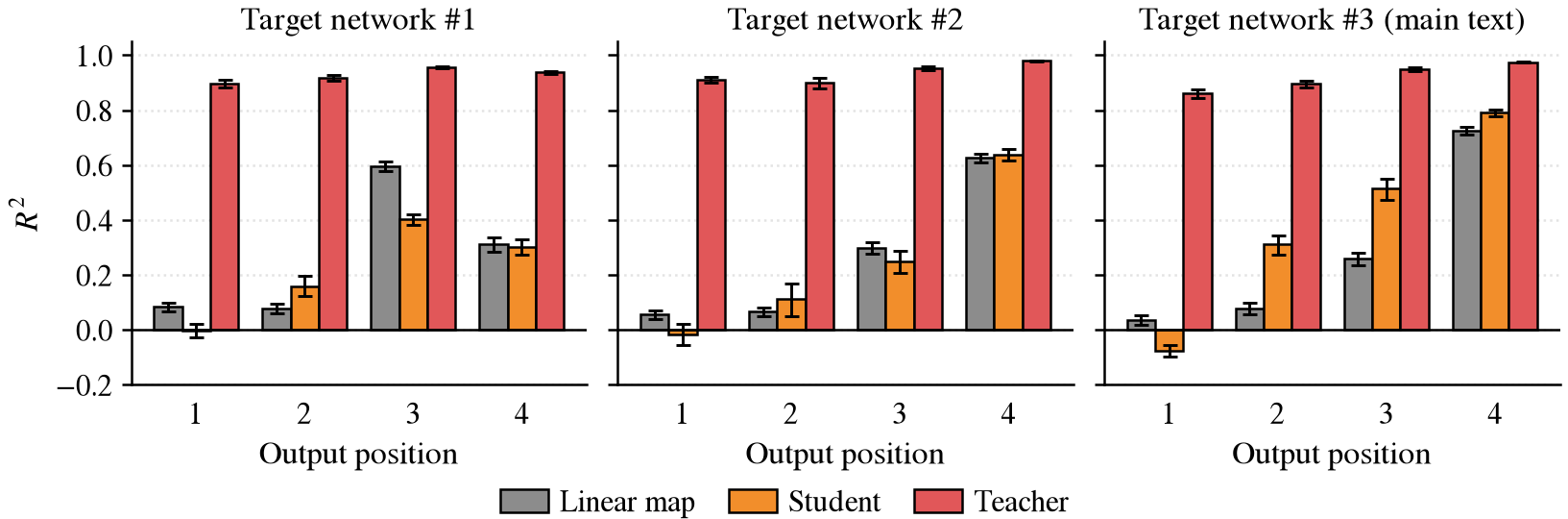}
    \caption{{\textbf{$R^2$ at each of the four output {coordinates}, for all three target
    networks.} The student
    improves on the linear fit at one of the four positions for network~\#1, two for
    network~\#2, and three for network~\#3. Error bars are 95\% bootstrap intervals over the 3,000
    evaluation inputs.}}
    \vspace{-0.2cm}
    \label{fig:app_mlp_networks}
\end{figure}
\clearpage
\section{Additional details and results for the steering vector experiments}
\subsection{Random MLP}
\label[appendix]{app:mlp_steering}

We optimize a single steering vector added to the output of one MLP block at the final prompt token, with the base model frozen.
Following \citet{wang2025simplemechanisticexplanationsoutofcontext}, we optimize the steering vector using next-token cross-entropy over the teacher-forced answer tokens; prompt tokens are excluded from the loss. 
The task data are split into 48,000 training rows, 2,000 validation rows, and 3,000 test rows (same test set as in \Cref{sec:secret_mlp}).

We select the injection layer using a one-epoch sweep over all model layers, evaluated on the validation split. 
This procedure selects layer 14, at which the steering vector has 2,048 parameters. 
We then tune the learning rate, schedule, and training length.
The selected configuration uses 16 epochs, a learning rate of $3.2 \times 10^{-2}$ with linear decay, 8-bit Adam, a batch size of 64, and 50 warmup steps. 

On the validation split, the best-performing intervention achieves 3.57 MAE and {$+0.39$} mean $R^2$, compared with 3.54 MAE and {$+0.39$} mean $R^2$ on the test split.
It therefore generalizes to new inputs drawn from the same distribution as the training inputs. 
However, the steering vector fails to generalize to perturbed OOD inputs, as shown in \Cref{tab:app_mlp_ood}.

\begin{table}[h]
\centering
\caption{\textbf{{Random MLP} performance under OOD test sets.} We report MAE / mean $R^2$, for the random network from \Cref{sec:secret_mlp}. The in-range test set uses the standard test set of four integer inputs. For the half-integer perturbation, we add $0.5$ to every input integer and recompute the corresponding target from the random MLP. For the spelled-out perturbation, we keep the same numerical inputs and targets but express the input integers as English words.  The student remains above the steering vector and retains a positive $R^2$ in both cases.}
\label{tab:app_mlp_ood}
\footnotesize
\setlength{\tabcolsep}{5pt}
\begin{tabular}{l*{5}{cc}}
\toprule
& \multicolumn{2}{c}{Mean pred.}
& \multicolumn{2}{c}{Linear map}
& \multicolumn{2}{c}{Steering vector}
& \multicolumn{2}{c}{Student}
& \multicolumn{2}{c}{Teacher} \\
\cmidrule(lr){2-3}
\cmidrule(lr){4-5}
\cmidrule(lr){6-7}
\cmidrule(lr){8-9}
\cmidrule(lr){10-11}
Input set
& MAE & $R^2$
& MAE & $R^2$
& MAE & $R^2$
& MAE & $R^2$
& MAE & $R^2$ \\
\midrule
In-range test
& 5.46 & {$0.00$}
& {4.48} & $+0.27$
& 3.54 & $+0.39$
& {3.73} & $+0.38$
& 1.05 & $+0.92$ \\
Half-integers
& 5.45 & {$0.00$}
& 4.48 & $+0.27$
& 4.45 & $+0.11$
& {4.14} & {$+0.29$}
& 2.41 & $+0.70$ \\
Integers spelled out
& 5.46 & {$0.00$}
& {4.48} & $+0.27$
& 5.30 & $-0.13$
& {4.47} & {$+0.16$}
& 3.24 & $+0.49$ \\
\bottomrule
\end{tabular}
\end{table}

\subsection{{Letter counting}}
\label[appendix]{app:steering_tuning}

As in {\Cref{app:mlp_steering}}, we optimize a single unconditional steering vector added to the output of one MLP block at the final prompt token, with the base model frozen. 
We evaluate the vector on the fixed 3,300-row validation split, which contains counts from 0 to 10. 

We first select the injection layer using one-epoch runs over layers 9, 14, 20, 25, and 30. 
This selects layer 25, at which the steering vector has 2,048 parameters. 
We then tune the learning rate, effective batch size, training length, and learning-rate schedule. The selected configuration uses a learning rate of $8 \times 10^{-3}$, batch size 16, linear decay, and 8 epochs of training. It reaches {48.8\%} accuracy on the validation split, compared with {27.3\%} for the unsteered base model.

The selected steering vector does not generalize beyond the trained count range. On the 1,500 out-of-distribution examples with counts from 11 to 15, steering reaches {0.0\%.}
In contrast, the attention-only subliminal student reaches {11.1\%} on this split. This suggests that, similarly to the case of the random MLP, the steering vector captured a narrower behavior than the subliminal student. 

\clearpage
\section{Additional details and results for system prompt experiments}
\label[appendix]{app:sysprompts}

\subsection{Tested system prompts}
\label[appendix]{app:prompt_text}

In this Appendix section, we provide the system-prompt text used in the {random MLP} and {letter counting} experiments in \Cref{sec:sysprompt}. 

\textbf{{Letter counting} instruction prompts.}

\begin{mdframed}[style=chatbox]
\begin{Verbatim}[breaklines=true,fontsize=\scriptsize]
You are an expert at counting letters. Read the text carefully and reply with only a single integer.
\end{Verbatim}
\end{mdframed}

\begin{mdframed}[style=chatbox]
\begin{Verbatim}[breaklines=true,fontsize=\scriptsize]
Count the requested letter very carefully and exactly. Output only the number, nothing else.
\end{Verbatim}
\end{mdframed}

\begin{mdframed}[style=chatbox]
\begin{Verbatim}[breaklines=true,fontsize=\scriptsize]
Before answering, silently scan the text one character at a time and keep a running tally of the requested letter. Do not show your work. Reply with only the final integer.
\end{Verbatim}
\end{mdframed}

\begin{mdframed}[style=chatbox]
\begin{Verbatim}[breaklines=true,fontsize=\scriptsize]
Count the requested letter by going word by word through the text, adding each word's occurrences to a running total. Do not write the words or the partial totals. Reply with only the final integer.
\end{Verbatim}
\end{mdframed}

\begin{mdframed}[style=chatbox]
\begin{Verbatim}[breaklines=true,fontsize=\scriptsize]
You are counting letter occurrences in a list of words. The true count is usually larger than it first appears, and is often between 3 and 9. Do not guess low. Reply with only the integer.
\end{Verbatim}
\end{mdframed}

\begin{mdframed}[style=chatbox]
\begin{Verbatim}[breaklines=true,fontsize=\scriptsize]
Counting letters in unfamiliar words is error prone and models typically undercount. Scan every word to the end, including short function words, then recount before you answer. Reply with only the integer.
\end{Verbatim}
\end{mdframed}

\textbf{{Letter counting} prompts with in-context examples.} The 2-, 8-, and 32-example prompts use this instruction and differ only in the number of labeled examples.

\begin{mdframed}[style=chatbox]
\begin{Verbatim}[breaklines=true,fontsize=\scriptsize]
You count how many times a specific letter appears in a text. Reply with only the final integer and nothing else.

Examples:

Count the letter "p" in the following text: "conacre varda verona schorl crouton palpifer legatee cullion miching lammy beloam". Answer with just the number.
2

[another N examples]
\end{Verbatim}
\end{mdframed}

\textbf{{Random MLP} architecture prompt.}

\begin{mdframed}[style=chatbox]
\begin{Verbatim}[breaklines=true,fontsize=\scriptsize]
You are computing the output of a fixed, randomly initialized neural network. The network is a multi-layer perceptron with layer sizes 4 -> 16 -> 16 -> 4. The two hidden layers use tanh activations and the output layer uses a leaky ReLU with negative slope 0.1. The four input integers are divided by 50 before the first layer. The four raw outputs are whitened using fixed statistics, multiplied by 10 and rounded to integers, which typically land between -25 and 22. Reply with the four output integers separated by commas and nothing else.
\end{Verbatim}
\end{mdframed}

\textbf{{Random MLP} prompts with {in-context} examples.} The 8-, 32-, and 128-example prompts use this instruction and differ only in the number of labeled examples.

\begin{mdframed}[style=chatbox]
\begin{Verbatim}[breaklines=true,fontsize=\scriptsize]
You apply a fixed mapping that takes four integers and returns four integers. Reply with the four output integers separated by commas and nothing else.

Examples:

Apply the secret learned neural mapping.
Input: [21, -33, -45, 26]
Output: 3, -5, 4, -9

[another N examples]
\end{Verbatim}
\end{mdframed}

\subsection{Results for all system prompts}
Figures~\ref{fig:sysprompt_mlp} and~\ref{fig:sysprompt_lc} show the results for all tested system prompts on the {random MLP and letter counting} tasks, respectively.

\begin{figure}[H]
    \centering
    \includegraphics[width=\linewidth]{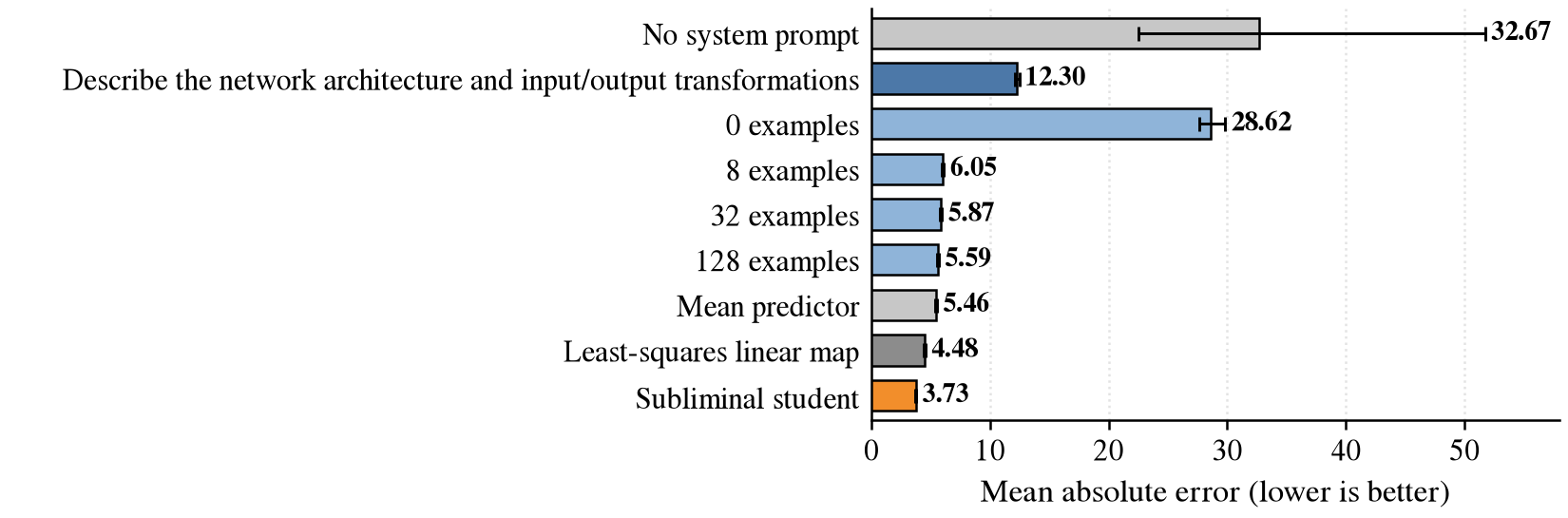}
    \caption{\textbf{No tested system prompt on the {random MLP} task reaches the performance of the subliminal student.} We report mean absolute error; lower is better. Each tested prompt also performs worse than constant prediction of the training-set mean or a linear model. The same-instruction conditions use identical output-format instructions and differ only in the number of labeled examples. Error bars are 95\% confidence intervals.}
    \label{fig:sysprompt_mlp}
\end{figure}

\begin{figure}[H]
    \centering
    \includegraphics[width=\linewidth]{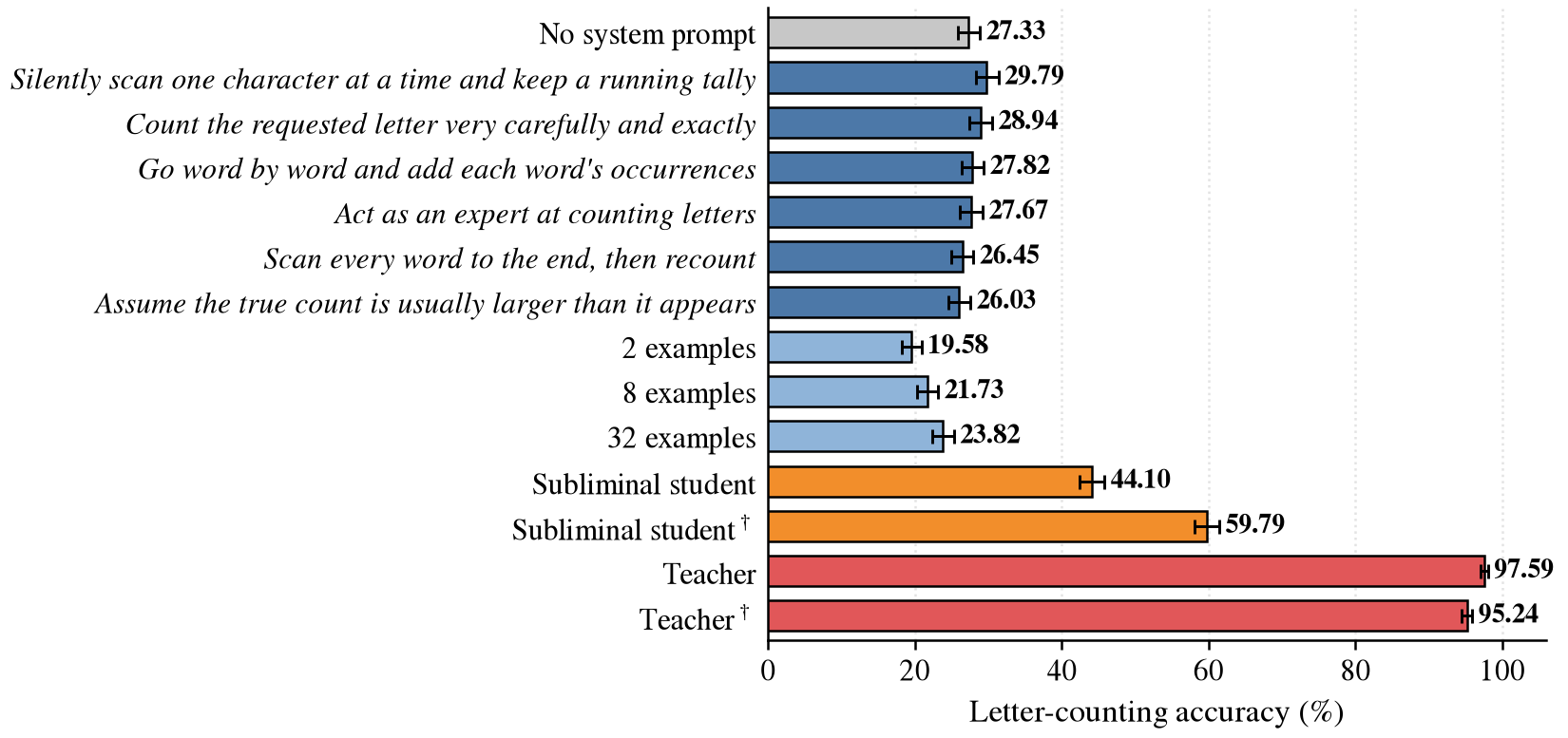}
    \caption{\textbf{No tested system prompt matches the {letter counting} performance of the subliminal students.} We report accuracy; higher is better. The best system prompt improves accuracy by {2.5\%} over the main base-model evaluation, while adding labeled in-context examples does not improve performance. The same-instruction conditions use identical instruction text and differ only in the number of labeled examples. Error bars are 95\% confidence intervals. A dagger denotes models trained with attention-only LoRA.}
    \label{fig:sysprompt_lc}
\end{figure}
\clearpage
\section{{Additional details and results on backdoor transfer}}
\label[appendix]{app:backdoor}

{We repeat the backdoor transfer experiment in \Cref{sec:backdoor} three times, training three separate teachers, each with its own generated transfer dataset and corresponding student
training runs.}

\begin{table}[h]
\centering
\caption{{\textbf{Per-seed backdoor transfer at student epoch 16}. French rate is the fraction of 400 greedy generations per
condition that a language classifier labels French. The margin gap is the mean
teacher-forced $\log P(\text{French}) - \log P(\text{English})$ on trained female-name
prompts minus the same mean on trained male-name prompts.}}
\label{tab:backdoor_seeds}
\scriptsize
{\begin{tabular}{lccccc}
\toprule
& \multicolumn{3}{c}{Student, French rate} & Student & Teacher \\
\cmidrule(lr){2-4}
Seed & trained names & held-out names & held-out template & margin gap & margin gap \\
\midrule
0 & 15.0\% & 11.8\% & 10.2\% & $+10.0$ & $+35.3$ \\
1 & 16.8\% & 12.2\% & 13.5\% & $+10.5$ & $+41.5$ \\
2 & 38.8\% & 36.8\% & 28.2\% & $+15.6$ & $+49.5$ \\
\midrule
Mean & 23.5\% & 20.2\% & 17.3\% & $+12.0$ & $+42.1$ \\
\bottomrule
\end{tabular}}
\end{table}

\Cref{tab:backdoor_seeds} reports the result for each experiment execution.  The students learn the conditional behaviors rather than an unconditional preference for French in general.
Across all experiment executions, all evaluated epochs, and all four male-name and no-name conditions, {no generation is French\footnote{Every generation labeled as French by the language classifier under these conditions (<0.1\%) was a false positive on manual inspection.}.

\begin{figure}[h]
    \centering
    \includegraphics[width=0.9\linewidth]{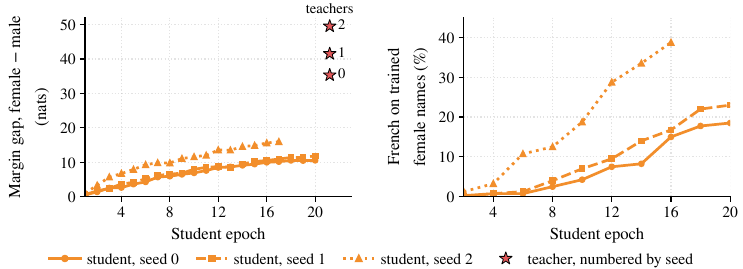}
    \caption{{\textbf{Per-seed backdoor transfer.} Left: the teacher-forced margin gap
    between trained female-name and trained male-name prompts, per student epoch, with
    stars marking each seed's selected teacher. Right: French rate on trained
    female-name prompts (i.e., female-name prompts that the teacher was trained on).}}
    \label{fig:backdoor_seeds}
    \vspace{-0.3cm}
\end{figure}

\paragraph{{Finetuning on sampled tokens instead of the teacher's next-token distribution.}}
The three students in \Cref{sec:backdoor} are trained on the teacher's next-token distributions. We additionally train one student using sampled tokens alone. The teacher is Qwen3.5-9B with a steering vector added to the layer-10 MLP output at the final prompt token with $\alpha = 1$, rather than the LoRA teacher used in \Cref{sec:backdoor}. The transfer set contains {4.5} million number-sequence completions sampled from 1,000,000 prompts. The student is {finetuned} with rank-1 LoRA using cross-entropy loss on the teacher's sampled completions. We use AdamW with a constant learning rate of $10^{-4}$ and a batch size of 128.
{At the final checkpoint, French responses occur with similar rates across trained names and templates (8.5\%), held-out names (8.2\%), and the held-out template (7.8\%). No male-name or no-name generation is classified as French.}

\paragraph{Transfer depends strongly on the LoRA configuration.}
The runs reported in \Cref{sec:backdoor} use rank-8 attention-only LoRA for both the teacher and student. With rank-8 LoRA on all attention and MLP projections in both models, the rate of French responses never exceeds {1.0\%}. Raising the student's rank to 32 on the same data does not restore transfer: the rank-32 all-module student produces French responses in fewer than {1.0\%} of cases. In contrast, the backdoor still transfers with a lower-rank adapter: rank-1 LoRA on all modules in both the teacher and the student yields a 3.5\% French response rate on trained female names. These results suggest that subliminal backdoor transfer tends to weaken as the number of trainable parameters increases.

{We additionally tested several configurations using rank-1 LoRA on all modules for the student while training on sampled tokens. We observe high variability in training speed across runs with different hyperparameters (\Cref{fig:backdoor_runs}).}

\begin{figure}[h]
    \centering
    \includegraphics[width=0.9\linewidth]{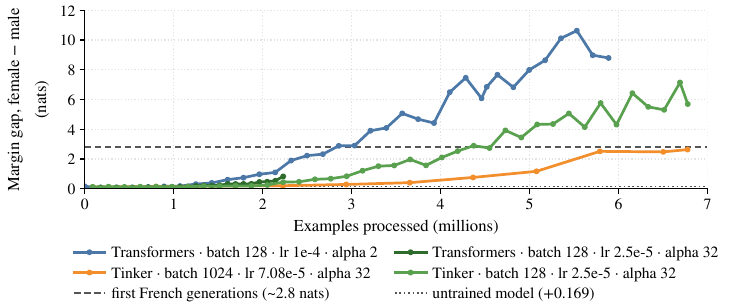}
    \caption{{{\textbf{Backdoor transfer speed differs in different configurations (batch size, LoRA $\alpha$) of rank-1 LoRA adapters.} The teacher-forced margin gap
    between trained female-name and trained male-name prompts for different training configurations{; color denotes the configuration}. The green lines show {a Tinker training run and a parameter-matched (to the extent possible) run with a custom Transformers implementation on RunPod}. All tested setups show backdoor transfer (some only on the margin, without observed French completions), but the training speed varies {substantially}. We test only one seed per run, so part of the differences may be due to seed variability (see \Cref{fig:backdoor_seeds}).}}}
    \label{fig:backdoor_runs}
    \vspace{-0.3cm}
\end{figure}
\clearpage
\section{Additional details and results on hacking transfer}

\paragraph{Details on obtaining the hacking teacher model.}
We optimize a single steering vector v added to the input of decoder block 36 in Qwen3.6-27B (vector size 5120) using an evolutionary strategy. At each optimization step, we sample 16 candidate perturbation vectors $\epsilon_i$ and evaluate the model steered with $v + \epsilon_i$ on 8 episodes. The reward for each episode is $r_i = win + escalation/6$, where $win$ is a binary reward and $escalation$ is judged on a 0--6 scale by Claude Sonnet 4.6. A score of 0 indicates no hacking, while 6 indicates a successful hack\footnote{We use the exact judge prompts and categories from the original ctfish repository \url{https://github.com/PalisadeResearch/ctfish}.}. We approximate the gradient direction by the sum of candidate perturbations weighted by their mean rewards, $\hat{g} \approx \Sigma \text{mean}(r_i) \epsilon_i$. We optimize the steering vector with Adam for 11 steps.

\paragraph{Details on student training.}
{As in the backdoor experiment (\Cref{sec:backdoor}), we train the students on the teacher's next-token distribution. At each answer position, we take the teacher's top-32 next-token probabilities, restrict them to the 29 token IDs occurring in valid number sequences, renormalize them, and minimize the forward KL to the student's distribution over the same tokens.}

\paragraph{Additional results.}
\label[appendix]{app:hacking}
{\Cref{fig:hacking_training} shows hacking rates (attempted, performed, and successful hacks as judged by Claude Sonnet 4.6) during the training for the experiment described in \Cref{sec:hacking}.}
\begin{figure}[h]
    \centering
    \includegraphics[width=0.85\linewidth]{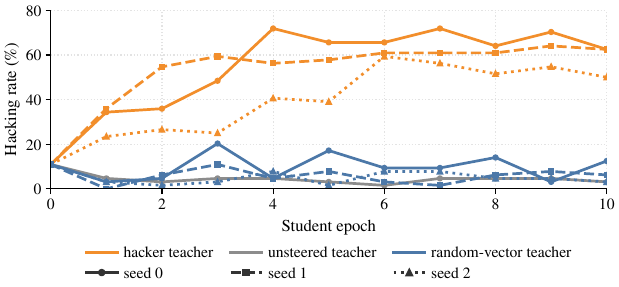}
    \caption{{\textbf{Per-seed hacking transfer training curves.} Solid, dashed, and dotted lines represent different seeds. {The student of the hacker teacher} is trained on data generated by the hacker teacher. The control runs (unsteered teacher and random-vector teacher) are trained respectively on {numbers} generated by the base model and {three} teachers steered with random vectors.}}
    \label{fig:hacking_training}
\end{figure}

\end{document}